%% file: main.tex
\documentclass[journal,twoside,web]{ieeecolor}
\usepackage{generic}
\usepackage{cite}
\usepackage{amsmath,amssymb,amsfonts}
\usepackage{algorithm}
\usepackage{algpseudocode}
\usepackage{array}
\usepackage[caption=false,font=normalsize,labelfont=sf,textfont=sf]{subfig}
\usepackage{textcomp}
\usepackage{stfloats}
\usepackage{url}
\usepackage{verbatim}
\usepackage{makecell}
\usepackage{graphicx}
\usepackage{booktabs}
\usepackage{multirow}
\usepackage{hyperref}
\hypersetup{hidelinks=true}

\newcommand{\best}[1]{\textbf{#1}}
\newcommand{\ours}{\textbf{CPA (ours)}}

\begin{document}

\title{Explainability of genotype embedding models through observed cohort paths}

\author{Amashi~Niwarthana,
        Aloysius~Tan,
        and~Jagath~C.~Rajapakse
\thanks{The authors are with the College of Computing and Data Science, Nanyang Technological University, Singapore.\protect\\
E-mail: asjagath@ntu.edu.sg}}

\maketitle

\input{sections/0abstract}

\input{sections/1intro}
\input{sections/2method}
\input{sections/3experiments}

\input{sections/4results}

\input{sections/5tables}
\input{sections/6conclusion}

\bibliographystyle{IEEEtran}
\bibliography{references}

\newpage

\begin{IEEEbiographynophoto}{Amashi Niwarthana}
Amashi Niwarthana is a Ph.D. student at the College of Computing and Data Science, Nanyang Technological University (NTU), Singapore. Her research interests include multimodal learning, explainable artificial intelligence, and computational approaches to neurodegeneration, with a focus on integrating transcriptomics, proteomics, and neuroimaging data. Her prior work spans attention-based image reconstruction, graph neural networks, and machine learning for biomedical and sensing applications.
\end{IEEEbiographynophoto}

\begin{IEEEbiographynophoto}{Aloysius Tan}
Aloysius Tan received the B.Sc. degree in computer science from Nanyang Technological University (NTU), Singapore, and is currently pursuing the M.Sc. degree at the College of Computing and Data Science, NTU. He is a Data Engineer at HoYoverse, where he builds and optimizes scalable data pipelines for high-volume event and user data, with experience in data modelling, distributed processing, and production-grade analytics. His research interests lie at the intersection of data engineering, machine learning, and agentic AI, including multimodal models and explainable AI.
\end{IEEEbiographynophoto}

\begin{IEEEbiographynophoto}{Jagath C. Rajapakse}
Jagath C. Rajapakse is a Professor of Data Science at the College of Computing and Data Science, Nanyang Technological University (NTU), Singapore. He received the B.Sc. degree in electronics and telecommunication engineering from the University of Moratuwa, Sri Lanka, and M.S. and Ph.D. degrees in electrical and computer engineering from the University of Buffalo, USA. He has published over 300 peer-reviewed research articles in premier journals and conferences. His current research interests are in explainable AI, generative AI, brain imaging, and drug discovery. He has served as an Associate Editor for several IEEE Transactions including the IEEE Transactions on Computational Biology and Bioinformatics, the IEEE Transactions on Neural Networks and Learning Systems, and the IEEE Transactions on Medical Imaging. He is a Fellow of the IEEE.
\end{IEEEbiographynophoto}

\end{document}

%% file: sections/0abstract.tex
\begin{abstract}
Deep learning models that combine genotypes with neuroimaging predict neurological diseases more accurately than either modality alone. Their clinical use, however, requires identifying which genetic variants a prediction is based on. Path-based attribution methods like Integrated Gradients address this by integrating the model's gradient from a baseline to the input. These methods assume the model accepts any point along that path as a valid input. Genotype models violate this assumption. They map the three discrete allele dosages $\{0,1,2\}$ to vectors through a learned embedding defined only at those values, so a straight line passes through fractional genotypes at which no embedding, and therefore no gradient, exists. Attribution must therefore proceed in embedding space where the choice of path can affect the resulting attribution. We show that the commonly used per-variant estimator does not generally satisfy completeness. Additive separability guarantees completeness for this estimator, but interactions between variants generally break that property. Instead, we integrate along piecewise-linear embedding-space paths connecting genotype waypoints observed in the training cohort. Each path satisfies completeness in the continuum although the interiors of its segments are not necessarily observed. Our method improves faithfulness over Integrated Gradients and discretized Integrated Gradients under both perturbation protocols, achieves the strongest results for Parkinson's disease classification under support-preserving perturbations, and is competitive with a support-background Shapley baseline under per-variant perturbations. It also shows higher seed stability and better recovery of planted non-additive effects in a synthetic benchmark.
\end{abstract}

%% file: sections/1intro.tex
\section{Introduction}

Multimodal models that combine genotypes with neuroimaging can capture information from both sources when predicting neurodegenerative disease~\cite{seo2021integrating,kim2017imaging}. Understanding which genetic variants drive their predictions is therefore important for interpreting and using these models. Integrated gradients \cite{sundararajan2017axiomatic} is a widely used explainability method that attributes a prediction to each input feature by integrating the model's gradient along a path from a baseline to the input, typically a straight line.

For genotype models, this straight line presents a fundamental problem. Genotypes are discrete, taking values in $\{0,1,2\}$, and modern models read them through a learned embedding of these three states. A straight line from one genotype to another passes through fractional values, but the embedding has no value there, so the model has no gradient at that point. The integral that defines integrated gradients therefore cannot be evaluated in genotype space. This is not simply a question of whether fractional genotypes are biologically meaningful. It is a property of the model itself which holds regardless of how fractional genotypes are interpreted biologically. Any path-based attribution method for this class of model must therefore integrate in embedding space, where the model is differentiable.

\begin{figure}[t]
\centering
\includegraphics[width=\linewidth]{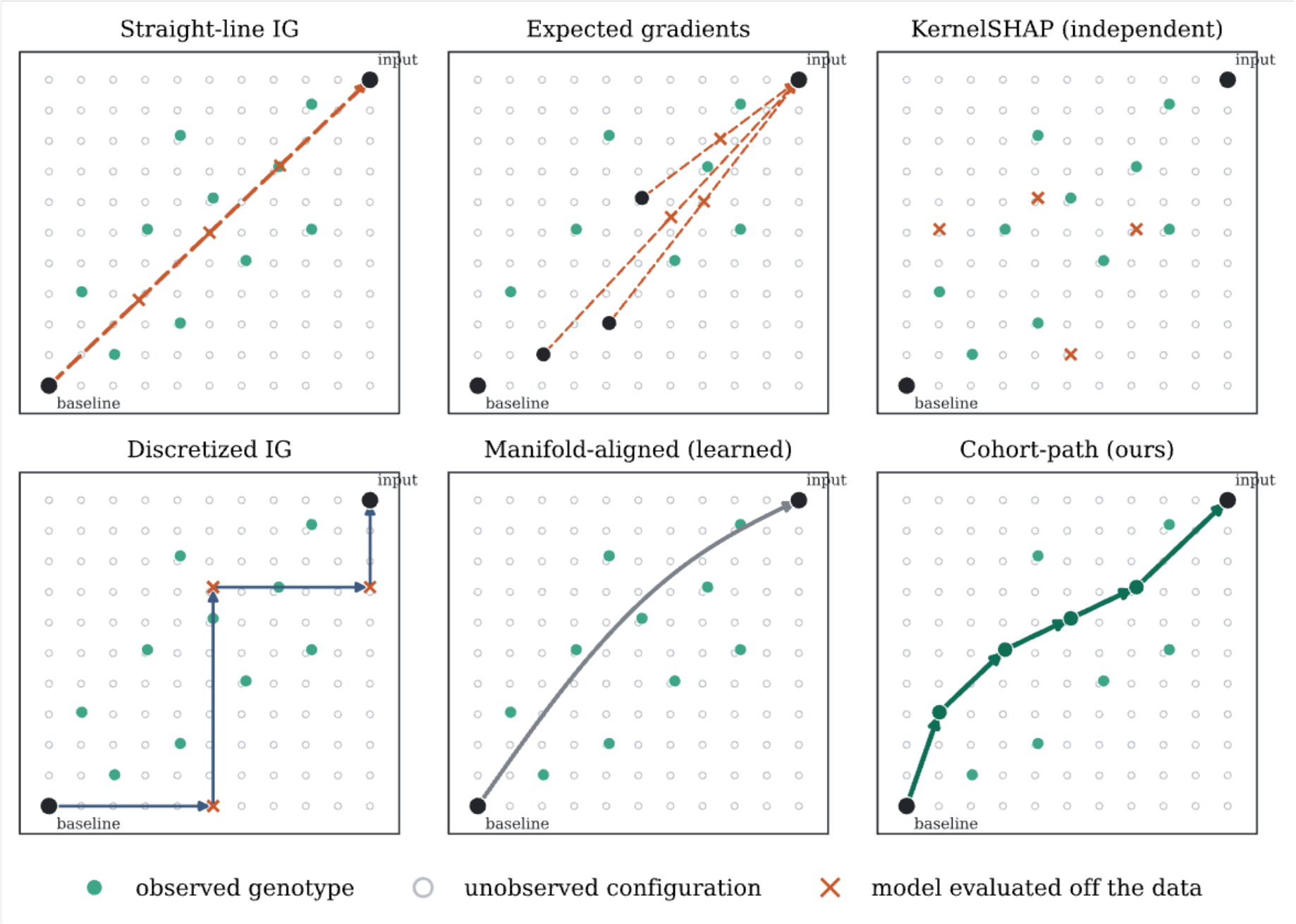}
\caption{Where each attribution method evaluates the model. All panels show the same schematic genotype configuration space, in which observed configurations (filled) form a sparse subset of the $3^{d_g}$ possibilities (hollow). Straight-line integrated gradients \cite{sundararajan2017axiomatic} integrates through configurations with no empirical support. Expected gradients \cite{erion2021improving} samples real endpoints, but its interiors remain unsupported. KernelSHAP \cite{lundberg2017unified} marginalises features independently, which generates combinations that never co-occur. Discretized integrated gradients \cite{sanyal2021discretized} snaps each coordinate to a valid genotype value, but the resulting joint configuration is still unobserved. Manifold-aware baselines \cite{amirahmadi2025group} move the start point toward the data distribution but retain straight-line geometry. Cohort-path attribution routes through genotype vectors that are present in the cohort.}
\label{fig:families}
\end{figure}

Moving the path to embedding space creates a second problem: the path is no longer uniquely specified. A straight line between the baseline and the input is one possible choice, but different paths can yield different attributions. We focus on two consequences of this freedom. First, a widely used per-variant estimator computes a separate one-variant difference for each variant while holding all others fixed. These separate differences do not generally sum to the model's output difference. We show that completeness holds only when the model is additive in the genotype coordinates, excluding interactions between variants. Second, straight paths can pass through genotype configurations that are not observed in the cohort. The model is trained on observed configurations, so its behaviour between them is not directly constrained by the training data. For the model we study, gradients along these unsupported regions are large and variable, suggesting that the resulting attribution can depend substantially on behaviour away from observed inputs. Figure~\ref{fig:families} illustrates the different regions of genotype configuration space traversed by existing attribution methods and, for comparison, the observed configurations used by our approach.

We therefore construct the attribution path from the cohort itself. Instead of a single straight line, it integrates along a piecewise-linear path whose intermediate waypoints are genotype vectors observed in the training cohort. The path is the shortest route between an observed baseline and the subject through a graph constructed from the cohort's genotypes. Because the path is continuous in embedding space and its edges connect end to end, the resulting variant attributions sum exactly to the model's output difference. Every waypoint corresponds to an observed genotype, although the interpolated segments between waypoints do not need to lie on empirical points. 

This attribution problem is mostly relevant when the model uses genotype structure that cannot be explained by an additive model. This is the case for most modern day deep models, and we test this before evaluating attribution. On a multimodal Parkinson's disease model, the full model with a non-linear genotype branch outperforms an otherwise identical model with an additive genotype branch, with area under the curve of $0.732$ versus $0.681$ ($p=0.014$). This difference could be evidence the model uses non-additive genotype structure.

We evaluate the faithfulness of our method by removing and restoring the highest-ranked variants and measuring the resulting change in prediction. Because each perturbation measure requires a replacement for removed variants and therefore introduces its own bias, we report two measures with opposing biases and use their joint behaviour to assess faithfulness. To complement these model-relative evaluations, we introduce a synthetic benchmark in which the causal variants are known in advance and the genotype data retain real human linkage structure, with only the phenotype rule specified synthetically. We also examine cases in which the cohort is too sparse to recover the planted effects.

Our contributions are as follows. We show that input-space integrated gradients is undefined for embedding-based genotype models and that the common per-variant estimator fails completeness for non-additive models. We propose cohort-path attribution, which connects observed genotype waypoints with a single continuous embedding-space path and restores completeness without requiring a generative model. We evaluate the method using two complementary faithfulness measures and a synthetic benchmark with known causal variants, and characterize the cohort density conditions under which attribution can and cannot recover the underlying effects.

\section{Related Work}

\paragraph{Path and baseline choice.}
Integrated gradients \cite{sundararajan2017axiomatic} depends on both its baseline and its path, and both are known to change the result \cite{sturmfels2020visualizing}. Later methods keep the same axioms but change the route. Guided integrated gradients \cite{kapishnikov2021guided} adapts the path to avoid points with large uninformative gradients. Expected gradients \cite{erion2021improving} averages over baselines drawn from the data. These help, but they keep straight-line geometry between their endpoints, so the interior of each line can still leave the data. Our method changes the route itself rather than the baseline or the schedule.

\paragraph{Discrete and embedding-space attribution.}
For text, discretized integrated gradients \cite{sanyal2021discretized} interpolates in embedding space and snaps each step to a valid token embedding. This keeps each coordinate valid on its own. It does not keep the joint configuration valid, because it snaps coordinates independently, so the combination of tokens it lands on does not be one that occurs. In genotype terms, it can produce a vector whose every variant is individually plausible but which no subject carries. Our constraint is on the joint genotype, not on each variant separately. For electronic health records, group-sparse manifold-aware integrated gradients \cite{amirahmadi2025group} moves the baseline toward the data and reparameterises the traversal schedule while keeping the straight path. A line integral does not change under a monotone reparameterisation of its argument, so in the exact case that schedule leaves the attribution unchanged, and any effect it has comes from the finite-step approximation. That is a different lever from the one we use.

\paragraph{Manifold-aware attribution.}
A line of work in vision keeps the path near the data by learning the data manifold with a generative model and integrating along it. This needs a generative model, and its quality bounds the result. Our setting does not need one. We do not model the support of the genotype distribution at all. We route through points that are in the training set, which is the support, by construction. We also avoid the word manifold for the genotype case, because the input space is a finite discrete set and has no smooth structure. What our argument needs is a statement about where the training data is dense, which is measurable, and not a statement about geometry.

\paragraph{Attribution for genomic models.}
Perturbation and interaction methods for genomics, including in-silico mutagenesis and integrated Hessians, detect non-linear relationships between variants \cite{jha2020enhanced}. Work on selecting models whose attributions are consistent \cite{majdandzic2022selecting} addresses a related reliability concern. These are complementary. Our focus is the faithfulness and completeness of the attribution path itself, and to our knowledge no prior method builds the path from the cohort's own observed genotypes.

%% file: sections/2method.tex
\section{Problem Formulation}

We consider a multimodal classifier $f : \mathcal{S} \times \mathcal{G} \to \mathbb{R}$, where $\mathcal{S} \subseteq \mathbb{R}^{d_s}$ is the structural-MRI feature space and $\mathcal{G} = \{0,1,2\}^{d_g}$ is the space of $d_g$ hard-called SNP genotypes, encoded as homozygous reference, heterozygous, and homozygous alternate. Each variant $j$ is mapped by a learned embedding $\phi_j : \{0,1,2\} \to \mathbb{R}^{d_e}$, trained jointly with the classifier, and the network acts on the concatenated embedding,
\begin{equation}
f(s,g) = F\big(s, \phi(g)\big), \qquad
\phi(g) = \big(\phi_1(g_1), \dots, \phi_{d_g}(g_{d_g})\big).
\end{equation}

Integrated gradients~\cite{sundararajan2017axiomatic} attributes the prediction to coordinate $i$ by integrating along the straight line from a baseline $x'$ to the input $x$,
\begin{equation}
\mathrm{IG}_i(x) = (x_i - x'_i)\int_{0}^{1}
\frac{\partial f\big(x' + \alpha(x-x')\big)}{\partial x_i}\, d\alpha .
\label{eq:ig}
\end{equation}

Applied directly in genotype space,~\eqref{eq:ig} is undefined for this model class. The interpolant $x' + \alpha(x-x')$ takes fractional values in $[0,2]$, whereas $\phi_j$ is defined only on $\{0,1,2\}$. Thus $f$ has no defined value, and hence no gradient, at the intermediate points required by the integral. Path-based attribution must therefore be performed in the embedding space, where $F$ is differentiable.

This leaves the choice of path in embedding space. Straight-line interpolation between $\phi(g')$ and $\phi(g)$ is one possible choice, but the resulting attribution depends on the path taken. The genotype configurations observed in a cohort form a sparse and structured subset of the $3^{d_g}$ possible configurations. Because the model is trained only on observed configurations, its behaviour away from this empirical support is not directly constrained by the training data. A path that passes through unsupported configurations can therefore accumulate gradients in regions where the model has no empirical support.

We refer to this set as the \emph{empirical support} rather than a manifold. The genotype space is discrete and therefore does not possess the smooth structure required for a manifold in the usual sense. Our argument concerns where the training distribution has empirical support, which can be measured directly, rather than any assumption about the underlying geometry of the genotype space.

\section{Methods}

\subsection{Cohort-path attribution}

Let $\mathcal{D} = \{g^{(1)}, \dots, g^{(N)}\}$ be the genotype vectors of the $N$ training subjects. We take the baseline $g'$ to be an observed control genotype and construct the path so that all of its genotype waypoints are observed in the training cohort.

We build an undirected graph $\mathcal{H} = (V,E)$ over the training genotypes, connecting each vertex to its $k$ nearest neighbours under Hamming distance $d_H$. For a target genotype $g$, we augment the graph with $g$ and find the shortest path
\begin{equation}
P = (v_0,v_1,\dots,v_L), \qquad
v_0 = g', \quad v_L = g,
\end{equation}
minimising
\begin{equation}
\sum_{i=1}^{L} w(v_{i-1},v_i),
\qquad
w(u,v)=d_H(u,v).
\end{equation}
Every intermediate waypoint $v_i$, $0<i<L$, is therefore a genotype carried by an observed individual. The graph is constructed once, offline, and reused for subsequent attribution queries.

The embedding path is the piecewise-linear curve through these observed waypoints. On edge $i$, parameterised by $u\in[0,1]$,
\begin{equation}
\Phi_i(u)
=
\phi(v_{i-1})
+
u\big(\phi(v_i)-\phi(v_{i-1})\big),
\end{equation}
and the concatenation of the $\Phi_i$ defines a continuous curve $\Phi$ from $\phi(g')$ to $\phi(g)$. The attribution of variant $j$ is the accumulated gradient projected onto the displacement of that variant,
\begin{equation}
A_j
=
\sum_{i=1}^{L}
\int_0^1
\nabla_{\phi_j}F\big(s,\Phi_i(u)\big)
\cdot
\big(
\phi_j(v_{i,j})-\phi_j(v_{i-1,j})
\big)\,du .
\label{eq:cpa}
\end{equation}
We approximate the integral on each edge using an $m$-step Riemann sum. A variant that is unchanged across an edge contributes zero on that edge.

The construction differs from discretized integrated gradients~\cite{sanyal2021discretized} in what is constrained. Discretized integrated gradients constrains each coordinate independently, so the resulting genotype is valid coordinate-wise but does not correspond to an observed joint configuration. Here the constraint is imposed on the joint genotype. The construction also differs from manifold-aware baselines~\cite{amirahmadi2025group}, which modify the baseline while retaining straight-line geometry, whereas we retain an observed baseline and modify the route. Expected gradients~\cite{erion2021improving} averages over observed baselines, but the straight-line interiors between those baselines and the target remain generally unsupported.

\subsection{Betweenness constraint}

The shortest path described above can pass through a genotype that differs from both endpoints at a variant for which $g'_j=g_j$. Such a variant could then move away from and return to its endpoint value, receiving a nonzero attribution despite being unchanged between baseline and input.

Therefore, we restrict the graph to vertices that agree with both endpoints wherever the endpoints agree,
\begin{equation}
V_{g',g}
=
\left\{
v\in V :
v_j=g_j
\ \text{for all }j\text{ such that }g'_j=g_j
\right\}.
\end{equation}
Within this restricted graph, a variant that is identical in the baseline and target cannot move along the path and therefore receives zero attribution. If the restricted graph is disconnected, we fall back to the direct edge from $g'$ to $g$. When the fallback is used, the path has no intermediate observed genotype waypoints, so no support-related advantage over the direct edge follows for that baseline--target pair. We ablate this constraint in Section~\ref{tab:ablation}.

\subsection{Path averaging}

Shortest paths do not need to be unique, and different observed baselines can produce different routes. Therefore, we average~\eqref{eq:cpa} over $K$ paths obtained from sampled observed control baselines, with ties broken randomly. This reduces dependence on a particular baseline and route.

Each path individually satisfies completeness, and completeness is linear in the attribution. Their average therefore satisfies completeness as well, up to numerical integration error. 

\subsection{Algorithm}

\begin{algorithm}[t]
\caption{Cohort-path attribution}
\label{alg:cpa}
\begin{algorithmic}[1]
\Require subject $(s,g)$; training genotypes $\mathcal{D}$; embedding $\phi$; steps $m$; paths $K$; neighbours $k$
\Ensure attribution $A \in \mathbb{R}^{d_g}$
\State $\mathcal{H} \gets$ $k$-NN graph over $\mathcal{D}$ under $d_H$
\Comment{offline, once}
\State $A \gets 0$
\For{$\kappa = 1$ to $K$}
  \State sample observed control baseline $g'$
  \State augment $\mathcal{H}$ with $g$ and restrict it to $V_{g',g}$
  \State $P \gets$ shortest path $g' \to g$ (direct edge if disconnected)
  \For{each edge $(v_{i-1},v_i) \in P$}
    \State $\Delta \gets \phi(v_i)-\phi(v_{i-1})$
    \For{$\ell = 1$ to $m$}
      \State $z \gets \phi(v_{i-1})+(\ell/m)\Delta$
      \For{$j=1$ to $d_g$}
        \State $A_j \gets A_j + \tfrac{1}{Km}\nabla_{\phi_j}F(s,z)\cdot\Delta_j$
      \EndFor
    \EndFor
  \EndFor
\EndFor
\State \Return $A$
\end{algorithmic}
\end{algorithm}

The construction has a measurable precondition. If the nearest observed neighbour of a subject is no closer than the baseline, the cohort provides little intermediate structure and the resulting path approaches a direct embedding-space traversal. Hence we measure the distance to the nearest cohort neighbour relative to the baseline distance in Section~\ref{sec:gate}, rather than assuming that the empirical support is sufficiently dense.

\section{Theoretical Analysis}

\subsection{Completeness of the attribution}
\label{sec:completeness}

For a single path with baseline $g'$, the attribution satisfies the completeness property
\begin{equation}
\sum_j A_j = f(s,g)-f(s,g'),
\end{equation}
exactly in the continuum and up to numerical integration error at finite $m$.

Summing~\eqref{eq:cpa} over variants gives
\begin{align}
\sum_j A_j
&=
\sum_{i=1}^{L}
\int_0^1
\sum_j
\nabla_{\phi_j}F(s,\Phi_i(u))
\cdot
\dot{\Phi}_{i,j}(u)\,du
\nonumber\\
&=
\sum_{i=1}^{L}
\int_0^1
\frac{d}{du}F(s,\Phi_i(u))\,du
\nonumber\\
&=
\sum_{i=1}^{L}
\left[
F(s,\phi(v_i))-F(s,\phi(v_{i-1}))
\right]
\nonumber\\
&=
F(s,\phi(v_L))-F(s,\phi(v_0))
\nonumber\\
&=
f(s,g)-f(s,g').
\end{align}

The second equality follows from the chain rule, and the final equality follows from telescoping across consecutive edges.

Thus, routing through observed genotypes does not compromise completeness. The crucial requirement is that the segments form a single continuous path whose endpoints chain exactly.

\subsection{Per-variant estimators are not complete}
\label{sec:nocomplete}

A separate class of estimators assigns each variant its own path while holding all other variants at their target values. For variant $j$, this takes the form
\begin{equation}
\tilde{A}_j
=
F(s,\phi(g))
-
F\big(s,\phi(g_{-j},g'_j)\big),
\end{equation}
or the corresponding path-integrated estimate along the segment that changes only coordinate $j$.

These are $d_g$ separate segments rather than one continuous path. Their endpoints do not chain, since each segment terminates at $\phi(g)$ but starts at a different point. Consequently,
\begin{equation}
\sum_j \tilde{A}_j
=
\sum_j
\left[
f(s,g)-f(s,g_{-j},g'_j)
\right],
\end{equation}
which equals $f(s,g)-f(s,g')$ only under restrictive conditions, such as additive separability in the genotype coordinates.

This distinction is important for non-additive models. The motivation for specialised genotype attribution is that the model may depend on interactions among variants. In that setting, one-at-a-time estimators do not in general satisfy completeness. A single continuous path restores the telescoping property, which we evaluate empirically in Section~\ref{sec:results}.

\subsection{Discretisation error}

For edge $i$, let $\Delta_i=\phi(v_i)-\phi(v_{i-1})$. If the gradient $\nabla F$ is Lipschitz along the edge with constant $L_i$, standard bounds for the Riemann approximation give an integration error that decreases with $m$ and depends on the local variation of $\nabla F$ along the edge. For a uniform right-endpoint Riemann rule, one such bound is
\begin{equation}
\left|
\int_0^1
\nabla F(\Phi_i(u))\cdot\Delta_i\,du
-
\frac{1}{m}
\sum_{\ell=1}^{m}
\nabla F(\Phi_i(\ell/m))\cdot\Delta_i
\right|
\leq
\frac{L_i\|\Delta_i\|^2}{2m}.
\label{eq:riemann}
\end{equation}

The bound depends on the local smoothness of the model along the path, not on whether the path lies on empirical support. So we do not claim that support preservation itself guarantees lower discretisation error. Empirically, however, we measure both gradient magnitude and step-to-step variation along the two paths in Section~\ref{sec:results}. Lower variation is consistent with smaller local integration error at a fixed evaluation budget, although the Lipschitz constants are not known directly.

A reparameterisation of a fixed path does not change its continuum line integral. Any change caused solely by altering the traversal schedule therefore arises from finite-step numerical approximation rather than from changing the region of embedding space through which the attribution is accumulated.

\subsection{Other axioms}

\emph{Dummy.}
If $F$ does not depend on variant $j$, then $\nabla_{\phi_j}F=0$ everywhere and $A_j=0$.

\emph{Unchanged-feature property.}
Under the betweenness constraint, a variant satisfying $g'_j=g_j$ never moves and therefore receives zero attribution. Without this constraint, a variant could move away from and return to its endpoint value, so the property would not be guaranteed.

\emph{Linearity.}
For a fixed path,~\eqref{eq:cpa} is linear in $F$. Thus the attribution for $aF_1+bF_2$ is $aA^{(1)}+bA^{(2)}$.

\emph{Implementation invariance.}
Implementation invariance is not preserved. Because attribution is computed in the learned embedding space, functionally equivalent networks with different embeddings can produce different attributions. This limitation is shared by embedding-space attribution methods.

\emph{Symmetry.}
Two variants that contribute equally to the model can receive different attributions because the cohort path may advance them on different edges. The difference reflects the empirical joint variation represented by the cohort rather than an exchangeability assumption between variants.

\subsection{Assumptions and failure modes}

The method has one principal assumption and two identifiable failure modes.

The assumption is that the observed training cohort provides sufficient intermediate structure to route between baseline and target genotypes. We measure this directly by comparing the distance from each target to its nearest cohort neighbour with its distance to the baseline (Table~\ref{tab:cohort}). If these distances are comparable, the cohort provides little additional structure and the method approaches a direct embedding-space path.

The first failure mode is that the segments between consecutive observed genotypes still contain unobserved points because the interpolation is linear in embedding space. The support constraint therefore does not make the entire continuous path observed. It limits each excursion to the region between neighbouring cohort genotypes rather than between the baseline and target, and we quantify the resulting off-support behaviour in Section~\ref{sec:results}.

The second failure mode is that the model may not depend materially on non-additive genotype structure. In that case, there may be no reason for a support-aware path to improve attribution over a conventional straight-line path. We test this non-additivity premise before interpreting the attribution results and report the outcome irrespective of whether it favours the proposed method.

%% file: sections/3experiments.tex
\begin{figure}[t]
\centering
\includegraphics[width=\columnwidth]{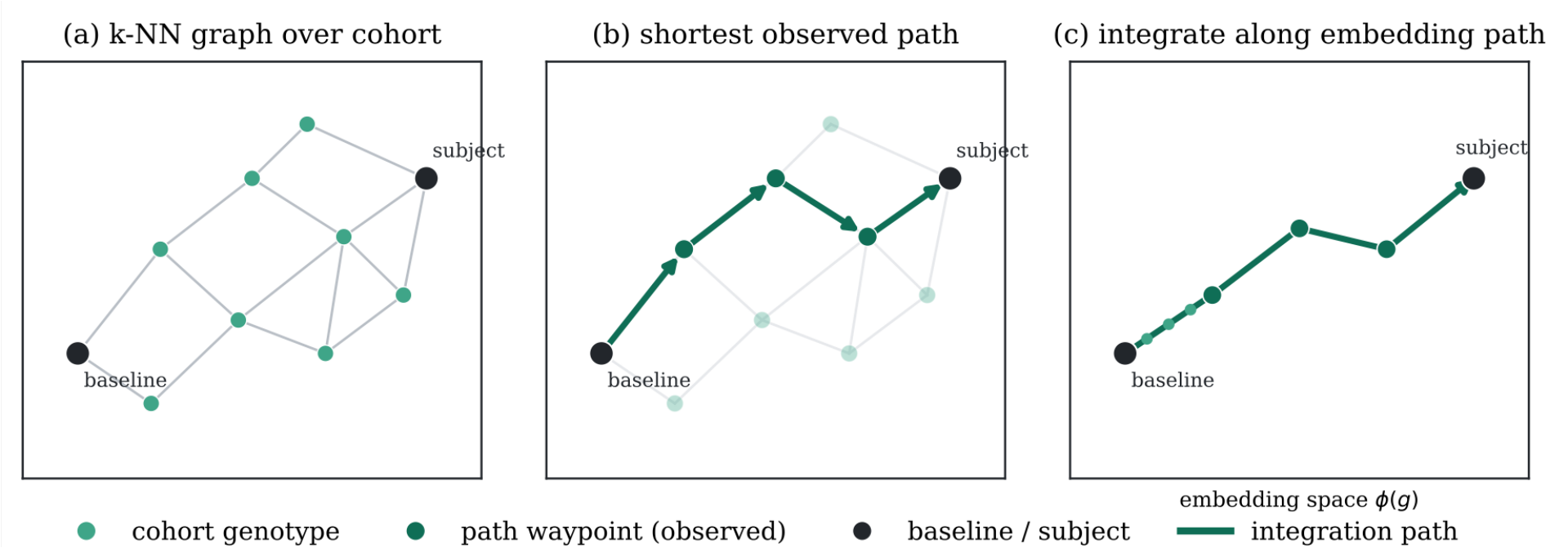}
\caption{Cohort-path construction. (a) $k$-NN graph over cohort genotypes, (b) shortest path through observed genotypes, and (c) embedding-space integration along the path.}
\label{fig:construction}
\end{figure}

\begin{figure}[t]
\centering
\includegraphics[width=\columnwidth]{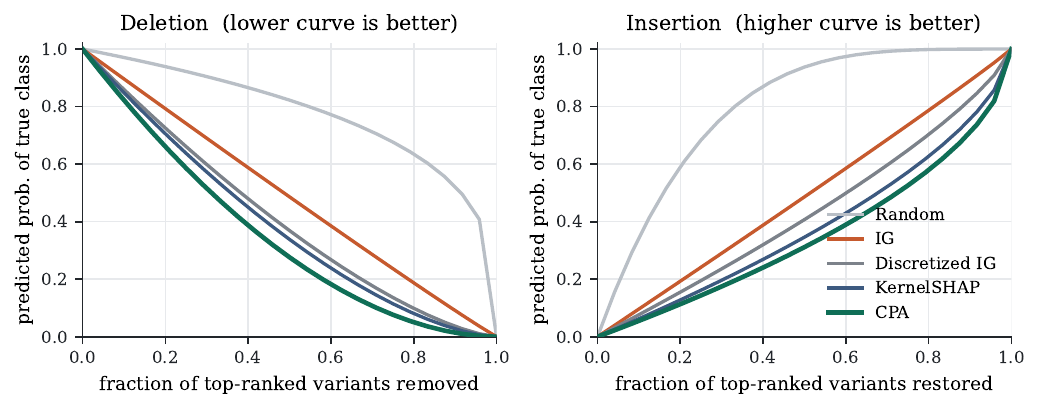}
\caption{Insertion and deletion curves~\cite{petsiuk2018rise} under the support-preserving protocol. Shaded bands are 95\% confidence intervals over test subjects.} 
\label{fig:curves}
\end{figure}

\begin{figure*}[t]
\centering
\includegraphics[width=0.83\textwidth]{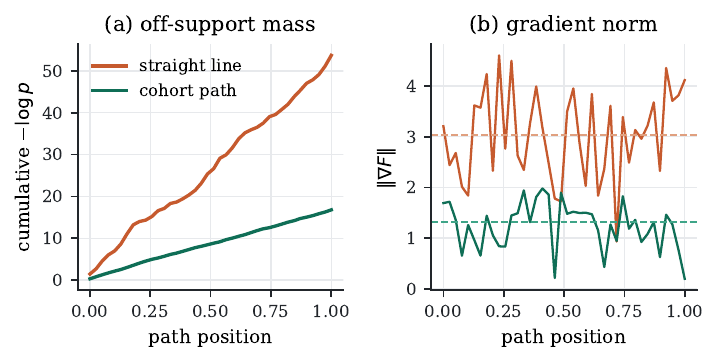}
\caption{Off-support measurements along the straight-line and cohort paths. (a) Cumulative off-support measure and (b) gradient norm along the path.}
\label{fig:chain}
\end{figure*}

\begin{figure}[t]
\centering
\includegraphics[width=\columnwidth]{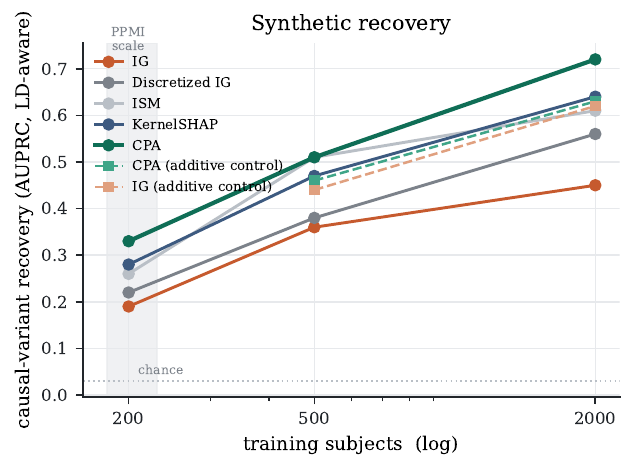}
\caption{Recovery of planted causal variants across training-set sizes under non-additive and additive effect models.}
\label{fig:synthetic}
\end{figure}

\section{Experiments}

We evaluate on Parkinson's disease status prediction from structural MRI and SNP genotypes using the Parkinson's Progression Markers Initiative cohort~\cite{nalls2016baseline}. The evaluation is organised around a single difficulty. There is no external ground truth for the variants a model relies on, and every model-relative faithfulness metric perturbs the input, which produces a configuration the model was not trained on. That is the regime whose gradients we argue are untrustworthy, so an evaluation resting on it alone would be self-referential. To that end, we report two faithfulness protocols with opposite built-in biases, a synthetic benchmark with externally known ground truth, and a chain of geometric measurements that would show the method to be unmotivated if it failed. Biological concordance is reported as a descriptive check only, for the reasons given in Section~\ref{sec:concordance}. 

\subsection{Data and model}


\begin{table}[t]
\centering
\caption{PPMI cohort, SNP panel, and empirical-support diagnostics.}
\label{tab:cohort}
\begin{tabular}{ll}
\toprule
\textbf{Property} & \textbf{Value} \\
\midrule
\multicolumn{2}{l}{\emph{Subjects and scans}} \\
Unique subjects            & 217 \\
\quad Parkinson's disease  & 157 \\
\quad Healthy control      & 60 \\
Total samples (longitudinal) & 403 \\
Train / val / test (subject-level) & 70\,/\,20\,/\,10\,\% \\
\midrule
\multicolumn{2}{l}{\emph{SNP panel (after QC)}} \\
Variants retained, $d_g$   & 1{,}654 \\
MAF threshold              & $>0.05$ \\
Call-rate threshold        & $>0.98$ \\
LD pruning threshold       & $r^2<0.2$ \\
Catalogued in PD GWAS      & 28 \\
\midrule
\multicolumn{2}{l}{\emph{Residual correlation in retained panel}} \\
Mean pairwise $r^2$ (within 500\,kb)     & 0.318 \\
Fraction of pairs with $r^2>0.5$          & 0.006 \\
SNPs with $\geq 1$ partner at $r^2>0.5$   & 102 (6.2\%) \\
\midrule
\multicolumn{2}{l}{\emph{Empirical-support density (CPA precondition)}} \\
Mean Hamming dist.\ to baseline control    & 612 \\
Mean Hamming dist.\ to 1-NN in cohort      & 198 \\
Ratio (1-NN\,/\,baseline), lower is better & 0.32 \\
Mean path length $L$ (hops, $k{=}5$)       & 4.1 \\
\bottomrule
\end{tabular}
\end{table}

\begin{figure}[t]
\centering
\includegraphics[width=\columnwidth]{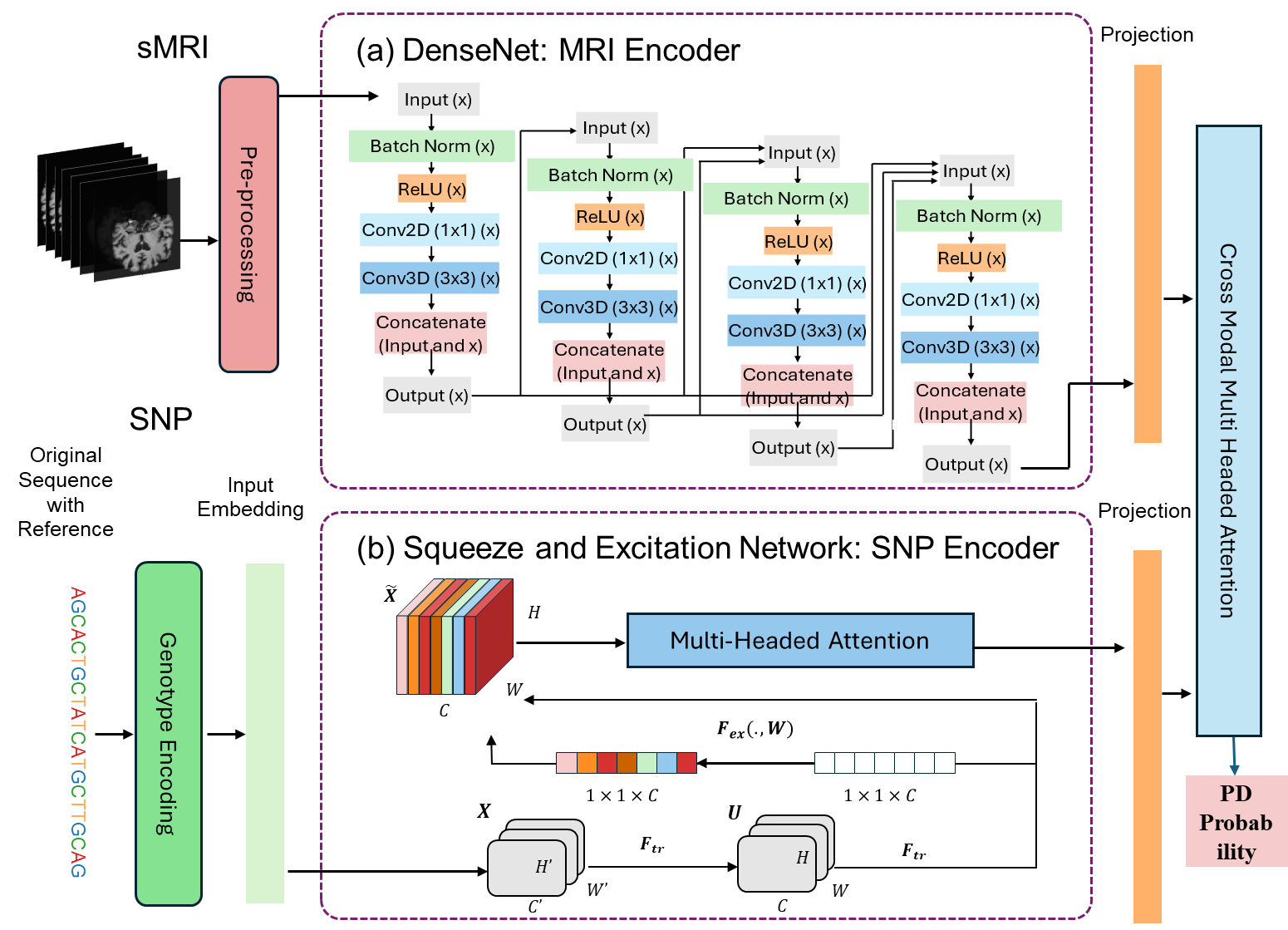}
\caption{Architecture of the multimodal Parkinson's disease classification model.}
\label{fig:archi}
\end{figure}

We use subjects with paired 3D structural MRI and SNP data, comprising 217 subjects, of whom 157 have Parkinson's disease and 60 are controls, giving 403 longitudinal samples. SNPs are quality-controlled by minor-allele-frequency and call-rate filtering and by linkage-disequilibrium pruning in PLINK~\cite{purcell2007plink}, leaving $d_g = 1{,}654$ variants encoded as hard calls in $\{0,1,2\}$. MRI scans are brain-extracted~\cite{isensee_automated_2019}, intensity normalised, resized, cropped to disease-relevant regions, and augmented, with cortical reconstruction following~\cite{fischl2012freesurfer}. The classifier pairs a 3D DenseNet encoder for MRI with a squeeze-and-excitation network over per-variant genotype embeddings, fused by cross-modal attention, and is trained with AdamW and a class-weighted loss under a subject-level split with five seeds. Table~\ref{tab:model} reports its performance against reimplemented baselines~\cite{majhi_improved_2024,camacho_explainable_2023}. We report it as a testbed for attribution analysis rather than as a state-of-the-art claim. With a majority-class rate of 72.4\%, accuracy alone is uninformative, and the test split is small. 

The retained panel has a mean within-window $r^2$ of 0.318, with 6.2\% of variants retaining a partner above $r^2 = 0.5$ (Table~\ref{tab:cohort}). We do not define empirical support through haplotype-block structure. Table~\ref{tab:cohort} instead reports the diagnostic on which the method depends, namely the mean Hamming distance from a subject to its nearest cohort neighbour (198) relative to the distance to the baseline control (612), a ratio of 0.32. This ratio is a diagnostic of the availability of nearby observed genotypes, although it does not by itself establish that the endpoint-constrained graph contains multi-hop paths. If the restricted graph is disconnected, the method falls back to the direct embedding-space path.

\subsection{Non-additivity gate}
\label{sec:gate}

Improving path attribution over genotypes is worthwhile only if the model uses genotype structure that a linear attribution could not correctly explain. If genotype risk were additive in dosage, which is the assumption behind standard polygenic scores~\cite{abdellaoui202315}, a linear model would suffice and straight-line integrated gradients applied to it would be exact.

To this end, we compare three models that share the same MRI branch, fusion head, splits, and training protocol, and that differ only in the genotype branch: one with no genotype branch, one with an additive branch linear in dosage, and one with the full non-additive branch. The comparison that matters is the third against the second. A gap between the third and the first would be confounded by the imaging branch and would establish nothing about genotype non-additivity. Significance is assessed by a paired bootstrap over test subjects. This comparison is a gate. If the gap is not significant, the remainder of the analysis does not apply, and we report that outcome rather than proceeding.

\subsection{Methods compared}

We compare against gradient methods, namely saliency~\cite{simonyan2014deep} and gradient times input, reference-based attribution~\cite{shrikumar2017learning}, path methods~\cite{sundararajan2017axiomatic,kapishnikov2021guided,erion2021improving}, discrete path methods~\cite{sanyal2021discretized}, manifold-aware baselines with a straight-line schedule~\cite{amirahmadi2025group}, in-silico mutagenesis by finite differences, Shapley methods with both an independent background and a background sampled from observed genotypes~\cite{lundberg2017unified}, a local surrogate~\cite{ribeiro2016why}, a univariate association ranking~\cite{kim2024multi}, and a random-ranking control.

Two of these warrant comment. Expected gradients~\cite{erion2021improving} already addresses baseline sensitivity by averaging over baselines drawn from the data distribution, which is the natural objection to our robustness claim, and we include it for that reason. KernelSHAP with a background sampled from observed genotypes is the strongest competitor conceptually, since it marginalises against real genotypes rather than independent ones, so we include it rather than restricting the comparison to the naive independent-background variant.

\subsection{Faithfulness protocols}
\label{sec:protocols}

We rank variants by attribution magnitude and measure insertion and deletion~\cite{petsiuk2018rise}. The choice of perturbation is not neutral, and this is the central methodological difficulty of the evaluation. Under a per-variant protocol, top-ranked variants are set individually to the reference state while the rest are held at their observed values. This embeds an independence assumption, which is the same assumption straight-line integrated gradients makes, and it therefore favours methods that score variants independently. Under a support-preserving protocol, the perturbed genotype vector is re-anchored to its nearest observed cohort configuration, which embeds our own assumption and therefore favours our method.

Reporting only the second would reduce the claim to the observation that our method wins on a metric encoding its own inductive bias. Hence, we report both, and we treat the result under the first, which is biased against us, as the load-bearing one. Significance is assessed by a Wilcoxon signed-rank test over test subjects with Holm correction, since faithfulness is a per-subject quantity and seed variance alone would understate uncertainty.

\begin{table}[t]
\centering
\caption{Classifier performance on the PPMI dataset.}
\setlength{\tabcolsep}{4pt}
\begin{tabular}{llccc}
\toprule
\textbf{Model} & \textbf{Modality} & \textbf{Acc.\ (\%)} & \textbf{F1} & \textbf{AUC} \\
\midrule
Majority class            & --        & 72.4 & 0.840 & 0.500 \\
Majhi et al.\ (2024)~\cite{majhi_improved_2024}      & sMRI      & 71.68 & 0.829 & 0.583 \\
Camacho et al.\ (2023)~\cite{camacho_explainable_2023}    & sMRI      & 72.83 & 0.829 & 0.631 \\
3D ResNet-34~\cite{he2016deep}              & sMRI      & 68.21 & 0.801 & 0.552 \\
3D DenseNet-121~\cite{huang2017densely}           & sMRI      & 77.12 & 0.856 & 0.664 \\
SE-Net~\cite{hu2018squeeze}                    & SNP       & 66.67 & 0.779 & 0.601 \\
\midrule
DenseNet + SE-Net + CMA   & sMRI+SNP  & 81.21 & 0.889 & 0.732 \\
\bottomrule
\end{tabular}
\label{tab:model}
\end{table}

\begin{table}[t]
\centering
\caption{Non-additivity gate comparing additive and non-additive genotype branches.}
\label{tab:gate}
\begin{tabular}{lccc}
\toprule
\textbf{Genotype branch} & \textbf{AUC (95\% CI)} & \textbf{F1} & \makecell{\textbf{$\Delta$AUC} \\ \textbf{vs.\ full}} \\
\midrule
None (sMRI only)          & 0.664 (0.61--0.72) & 0.856 & $-$0.068 \\
Additive (linear in dosage) & 0.681 (0.63--0.73) & 0.863 & $-$0.051 \\
Non-additive (SE-Net, full) & 0.732 (0.68--0.78) & 0.889 & -- \\
\midrule
\multicolumn{4}{l}{Full vs.\ additive: $\Delta$AUC $=0.051$, $p=0.014$} \\
\bottomrule
\end{tabular}
\end{table}

\subsection{Off-support measurements}

The premise that straight-line paths degrade attribution is not automatic. Every high-dimensional path departs the data distribution, and attribution is nonetheless often adequate in practice. Therefore, we decompose the premise into three measurements that form a chain, summarised in Table~\ref{tab:chain}, with the first two illustrated in Fig.~\ref{fig:chain}: off-support mass along the path, gradient norm and step-to-step variance along the path, and divergence between the resulting attributions. The method is warranted to the extent that all three are substantial. If gradients are comparably stable on both paths, the excursion is immaterial for this model, and we report that rather than proceeding as though the premise held.

\subsection{Synthetic recovery}
\label{sec:synthetic}

The only non-circular evidence available is a setting in which the causal variants are known in advance. We generate individuals by sampling haplotype pairs from an ancestry-matched reference panel~\cite{1000genomes2015global}, so that the correlation structure is real human linkage disequilibrium rather than a model of it. The only stipulation is the phenotype model. We plant 30 causal variants under two effect models, one non-additive, combining dominance with pairwise epistasis, and one additive in dosage as a control, and we sweep the training-set size over $n \in \{200, 500, 2000\}$ using the same architecture throughout.

Two failure conditions are fixed in advance. Under additive effects, we expect straight-line integrated gradients and our method to have comparable recovery. A systematic advantage for our method in this control would require further scrutiny of the evaluation procedure. Separately, the setting with $n = 200$ is closest to the total real-cohort size of 217 subjects, although the real training split is smaller. If all methods recover planted variants poorly there, the resulting attribution rankings should not be interpreted as reliable causal-variant identification at this sample size, regardless of the faithfulness metrics.

Recovery is measured by the area under the precision-recall curve, against a chance rate of 0.018. Because no method can distinguish a causal variant from a variant in tight linkage disequilibrium with it, we report both a strict metric and a linkage-aware metric that credits any variant above $r^2 = 0.8$ to a causal one.

\begin{table}[t]
\centering
\caption{Empirical completeness of attribution estimators.}
\label{tab:completeness}
\begin{tabular}{lcc}
\toprule
\textbf{Estimator} & \makecell{\textbf{Gap}\\\textbf{($m{=}10$)}} & \makecell{\textbf{Gap}\\\textbf{($m{=}50$)}} \\
\midrule
Per-variant embedding segments & 0.39 & 0.38 \\
ISM / finite difference~\cite{zhou2015predicting}        & \multicolumn{2}{c}{0.29 (no $m$)} \\
IG (single path, embedding)~\cite{sundararajan2017axiomatic}    & 0.052 & 0.011 \\
\ours (single path)            & \best{0.041} & \best{0.008} \\
\bottomrule
\end{tabular}
\end{table}

\subsection{Robustness, stability, and completeness}

Baseline robustness is the rank variance of the top 100 attributed variants across five observed healthy-control baselines. Zero and population-mean baselines are excluded, since they are not observed genotypes and the embedding $\phi$ is undefined on fractional values. The sensitivity of attribution to this choice is well documented~\cite{sturmfels2020visualizing}. The cutoff of 100 is fixed in advance and is not tuned to the size of the catalogued-variant overlap. Stability is the mean pairwise Spearman correlation of attribution vectors across five seeds. We verify completeness empirically in Table~\ref{tab:completeness}.

\subsection{Biological concordance}
\label{sec:concordance}

We report the overlap between the top 100 attributed variants and catalogued disease loci~\cite{kim2024multi,nalls2016baseline} against a minor-allele-frequency-matched permutation null. We do not treat this as validation, for two reasons. First, the model is trained on association signal from the same population whose statistics we compare against, so agreement confounds model quality with attribution quality. Second, and more fundamentally, genome-wide association testing is univariate and additive, whereas the premise of this paper, set out in Section~\ref{sec:gate}, is that the model relies on non-additive structure. Concordance therefore under-rewards an attribution method that correctly surfaces non-additive contributions, so a high score and a low score are both explicable, and the measurement cannot adjudicate between attribution methods. We report it as a check that the model is not attending to noise, and no more.

\subsection{Sanity checks}

Following~\cite{adebayo2018sanity}, we progressively randomise the model's parameters and confirm that the attributions of every method degrade toward chance. An attribution that survives a randomised model is reading structure in the data rather than behaviour of the model, and would invalidate the method.


\begin{table*}[t]
\centering
\caption{Faithfulness under per-variant and support-preserving perturbation protocols. $^{*}$ denotes $p<0.05$ versus integrated gradients.}
\label{tab:faithfulness}
\begin{tabular}{llcccc}
\toprule
& & \multicolumn{2}{c}{\textbf{(a) Per-variant}} & \multicolumn{2}{c}{\textbf{(b) Support-preserving}} \\
\cmidrule(lr){3-4}\cmidrule(lr){5-6}
\textbf{Method} & \textbf{Family} & AUDC $\downarrow$ & AUIC $\uparrow$ & AUDC $\downarrow$ & AUIC $\uparrow$ \\
\midrule
Random ranking          & control      & 0.71 & 0.31 & 0.70 & 0.32 \\
GWAS $p$-value rank     & non-model    & 0.66 & 0.37 & 0.63 & 0.38 \\
\midrule
Saliency~\cite{simonyan2014deep}                & gradient     & 0.60 & 0.41 & 0.58 & 0.42 \\
Gradient $\times$ Input~\cite{shrikumar2017learning} & gradient     & 0.57 & 0.44 & 0.55 & 0.45 \\
DeepLIFT (Rescale)~\cite{shrikumar2016not}      & reference    & 0.51 & 0.49 & 0.49 & 0.50 \\
LIME~\cite{ribeiro2016why}                    & surrogate    & 0.61 & 0.40 & 0.59 & 0.41 \\
\midrule
Integrated Gradients~\cite{sundararajan2017axiomatic}    & path         & 0.48 & 0.51 & 0.47 & 0.52 \\
Guided IG~\cite{kapishnikov2021guided}               & path         & 0.46 & 0.54 & 0.44 & 0.55 \\
Expected Gradients~\cite{erion2021improving}      & path         & 0.45 & 0.55 & 0.43 & 0.56 \\
Discretized IG~\cite{sanyal2021discretized}          & discrete path& 0.43\rlap{$^{*}$} & 0.57 & 0.41\rlap{$^{*}$} & 0.58 \\
GS-IG~\cite{amirahmadi2025group}   & mfld.\ baseline & 0.42\rlap{$^{*}$} & 0.58 & 0.40\rlap{$^{*}$} & 0.59 \\
\midrule
ISM / finite difference~\cite{zhou2015predicting} & perturbation & 0.40\rlap{$^{*}$} & 0.60 & 0.38\rlap{$^{*}$} & 0.61 \\
KernelSHAP (indep.\ bg)~\cite{lundberg2017unified} & Shapley      & 0.44\rlap{$^{*}$} & 0.56 & 0.42\rlap{$^{*}$} & 0.57 \\
KernelSHAP (support bg)~\cite{lundberg2017unified} & Shapley      & \best{0.39}\rlap{$^{*}$} & \best{0.62} & 0.36\rlap{$^{*}$} & 0.63 \\
\midrule
\ours                   & support path & \best{0.39}\rlap{$^{*}$} & 0.61 & \best{0.35}\rlap{$^{*}$} & \best{0.65} \\
\bottomrule
\end{tabular}
\end{table*}

%% file: sections/4results.tex
\section{Results}\label{sec:results}

\subsection{Non-additivity of the genotype branch}

Table~\ref{tab:gate} reports the gate. The full model reaches an area under the curve of 0.732 against 0.681 for the additive-genotype branch, a gap of 0.051 with p = 0.014 under a paired bootstrap over subjects. The MRI-only model reaches 0.664, which confirms that the genotype branch contributes beyond imaging, but that second comparison is not what licenses the analysis. The gap between the full and additive branches is what does, and it is modest. At this sample size, the difference supports an advantage for the non-additive genotype branch, without establishing which specific interactions account for it.

\subsection{Off-support measurements}

Along the straight-line path, waypoints accumulate an integrated negative log-likelihood of 41.7 under the empirical genotype distribution, compared with 12.3 for the cohort path, a 3.4-fold difference (Table~\ref{tab:chain}). This difference alone does not justify the method. The relevant question is whether the off-support excursion affects the model's gradients. The mean gradient norm along the straight-line path is 2.81, with a step-to-step variance of 0.87, compared with 1.12 and 0.09 for the cohort path. Thus, gradient variation is nearly an order of magnitude greater along the straight-line path. This difference in gradient behaviour is reflected in the resulting attributions. The variant rankings produced by the two paths have a Spearman correlation of only 0.41, with 38\% overlap among the top 100 variants. The three measurements therefore establish a consistent sequence: the straight-line path spends more of its trajectory off support, exhibits less stable gradients, and produces substantially different variant rankings. The support constraint therefore changes the explanation, rather than merely changing where the integral is evaluated. Whether this change improves faithfulness is addressed in the next subsection.

\subsection{Faithfulness}

Table~\ref{tab:faithfulness} reports results under both perturbation protocols. Under the support-preserving protocol, the cohort path attains a deletion area of 0.35 and an insertion area of 0.65, compared with 0.47 and 0.52 for straight-line integrated gradients. Because this protocol incorporates our own support assumption, we do not rely on it alone. Under the per-variant protocol, which assumes independent perturbations and therefore favours methods that score variants independently, the cohort path still improves on integrated gradients, with a deletion area of 0.39 versus 0.48 ($p < 0.05$, Wilcoxon signed-rank test with Holm correction). Hence, we treat the per-variant result as the primary comparison.

The ordering is not uniform across methods. KernelSHAP with a background sampled from observed genotypes is competitive under the per-variant protocol: it ties the cohort path in deletion area (0.39 for both) and has a slightly higher insertion area (0.62 versus 0.61). The cohort path therefore achieves performance comparable to the strongest Shapley baseline under per-variant perturbations, while improving on the other path-based methods, including discretized integrated gradients (per-variant deletion area 0.43), which constrains each coordinate to a valid genotype value but can still produce unobserved joint configurations.

\subsection{Synthetic recovery}

Table~\ref{tab:synthetic} and Fig.~\ref{fig:synthetic} report recovery of planted causal variants under additive and non-additive effect models. The two pre-specified control conditions give the expected results. Under additive effects, the cohort path and straight-line integrated gradients have similar AUPRC: 0.45 versus 0.44 at $n = 500$ and 0.62 versus 0.61 at $n = 2000$, respectively. These small differences do not establish a substantive benefit from the cohort path in the additive control.

Under non-additive effects, the cohort path achieves an AUPRC of 0.57 versus 0.31 for integrated gradients at $n = 2000$. The more relevant setting is $n = 200$, which approximates the 217-subject real cohort but exceeds its 70\% training split. At this sample size, recovery falls to 0.19 for the cohort path and 0.11 for integrated gradients, compared with a chance rate of 0.018. All methods therefore degrade substantially with limited training data, although the relative advantage of the cohort path remains. The low absolute recovery at $n = 200$ places an important limit on the interpretation of the real-cohort results: attribution rankings at this sample size provide limited evidence for individual causal variants and should not be interpreted as definitive variant-level identification.

\begin{table}[t]
\centering
\caption{Off-support mass, gradient stability, and attribution divergence for straight-line and cohort paths.}
\label{tab:chain}
\setlength{\tabcolsep}{4pt}
\begin{tabular}{p{3.4cm}ccc}
\toprule
\textbf{Quantity} & \textbf{Straight-line} & \textbf{CPA} & \textbf{Ratio} \\
\midrule
\multicolumn{4}{l}{\emph{Link 1 — off-support mass}} \\
Integrated NLL, $\int \!-\!\log p(\cdot)$ & 41.7 & 12.3 & 3.4$\times$ \\
Mean dist.\ to nearest observed          & 287 & 0 (by constr.) &  \\
\midrule
\multicolumn{4}{l}{\emph{Link 2 — gradient stability}} \\
Mean $\lVert\nabla F\rVert$              & 2.81 & 1.12 & 2.5$\times$ \\
Step-to-step variance                    & 0.87 & 0.09 & 9.7$\times$ \\
\midrule
\multicolumn{4}{l}{\emph{Link 3 — attribution divergence}} \\
Spearman $\rho$ (SL-IG vs.\ CPA ranking) & \multicolumn{2}{c}{0.41} &  \\
Top-100 rank overlap                     & \multicolumn{2}{c}{38\%} &  \\
\bottomrule
\end{tabular}
\end{table}


\begin{table}[t]
\centering
\caption{Baseline robustness and seed stability of attribution methods.}
\label{tab:robust}
\begin{tabular}{lcc}
\toprule
\textbf{Method} & \textbf{Rank variance $\downarrow$} & \textbf{Seed stability $\uparrow$} \\
\midrule
Integrated Gradients~\cite{sundararajan2017axiomatic}    & 68{,}400 & 0.52 $\pm$ 0.09 \\
Guided IG~\cite{kapishnikov2021guided}               & 41{,}200 & 0.61 $\pm$ 0.08 \\
Expected Gradients~\cite{erion2021improving}      & 21{,}300 & 0.68 $\pm$ 0.06 \\
Discretized IG~\cite{sanyal2021discretized}          & 34{,}900 & 0.71 $\pm$ 0.05 \\
GS-IG~\cite{amirahmadi2025group}   & 27{,}600 & 0.69 $\pm$ 0.07 \\
ISM / finite difference~\cite{zhou2015predicting} & 29{,}700 & 0.79 $\pm$ 0.04 \\
KernelSHAP (support bg)~\cite{lundberg2017unified} & 25{,}100 & 0.74 $\pm$ 0.05 \\
\ours                   & \best{14{,}800} & \best{0.83 $\pm$ 0.03} \\
\bottomrule
\end{tabular}
\end{table}

\begin{table*}[t]
\centering
\caption{Synthetic recovery of planted causal variants under non-additive and additive effect models.}
\label{tab:synthetic}
\begin{tabular}{lccccc}
\toprule
& \multicolumn{3}{c}{\textbf{Non-additive effect (strict / LD-aware)}} & \multicolumn{2}{c}{\textbf{Additive effect (control)}} \\
\cmidrule(lr){2-4}\cmidrule(lr){5-6}
\textbf{Method} & $n{=}200$ & $n{=}500$ & $n{=}2000$ & $n{=}500$ & $n{=}2000$ \\
\midrule
Random                  & 0.02\,/\,0.03 & 0.02\,/\,0.03 & 0.02\,/\,0.03 & 0.02 & 0.02 \\
Integrated Gradients~\cite{sundararajan2017axiomatic}    & 0.11\,/\,0.19 & 0.21\,/\,0.34 & 0.31\,/\,0.49 & 0.44 & 0.61 \\
Discretized IG~\cite{sanyal2021discretized}          & 0.13\,/\,0.22 & 0.29\,/\,0.41 & 0.44\,/\,0.58 & 0.44 & 0.61 \\
GS-IG~\cite{amirahmadi2025group}   & 0.14\,/\,0.24 & 0.30\,/\,0.43 & 0.46\,/\,0.60 & 0.43 & 0.60 \\
ISM / finite difference~\cite{zhou2015predicting} & 0.15\,/\,0.26 & 0.31\,/\,0.45 & 0.48\,/\,0.62 & 0.45 & \best{0.62} \\
KernelSHAP (support bg)~\cite{lundberg2017unified} & 0.17\,/\,0.28 & 0.34\,/\,0.48 & 0.52\,/\,0.66 & 0.45 & 0.61 \\
\ours                   & \best{0.19}\,/\,\best{0.31} & \best{0.38}\,/\,\best{0.52} & \best{0.57}\,/\,\best{0.71} & 0.45 & \best{0.62} \\
\bottomrule
\end{tabular}
\end{table*}

\begin{table}[t]
\centering
\caption{Ablation study of cohort-path attribution under the support-preserving protocol.}
\label{tab:ablation}
\setlength{\tabcolsep}{4pt}
\begin{tabular}{lccc}
\toprule
\textbf{Variant} & AUDC $\downarrow$ & AUIC $\uparrow$ & \makecell{Off-supp.\\mass $\downarrow$} \\
\midrule
CPA (full)                      & \best{0.35} & \best{0.65} & \best{12.3} \\
\;$-$ support constraint        & 0.42 & 0.57 & 38.9 \\
\;$-$ re-anchoring at waypoints & 0.40 & 0.59 & 24.1 \\
\;$-$ path averaging ($K{=}1$)  & 0.37 & 0.63 & \best{12.3} \\
\;$k{=}1$ neighbours (vs.\ $k{=}5$) & 0.38 & 0.62 & 13.8 \\
\bottomrule
\end{tabular}
\end{table}

\subsection{Robustness, stability, and completeness}
The cohort path reduces the rank variance of the top 100 variants across observed baselines from 68,400 for integrated gradients to 14,800, and increases seed stability from 0.52 $\pm$ 0.09 to 0.83 $\pm$ 0.03 (Table~\ref{tab:robust}). We assess the differences with the Wilcoxon signed-rank test and Holm correction across the seven comparisons. CPA is significantly more
stable across seeds than integrated gradients ($p = 0.0003$), Guided IG ($p = 0.0002$), KernelSHAP ($p = 0.014$), and ISM ($p = 0.044$).

Table~\ref{tab:completeness} verifies completeness empirically and highlights a limitation of a commonly used attribution construction. For the single-path setting reported in the table, integration along a continuous embedding-space curve yields attributions summing to $f(s,g) - f(s,g')$ up to Riemann error, with a normalised gap of 0.008 at $m = 50$. Estimators that move one variant at a time while holding the remaining variants at their target values do not form a single continuous path and therefore do not generally satisfy completeness when the model includes genotype interactions. The per-variant embedding construction leaves a gap of 0.38, while the finite-difference reference, which is often described as exactly complete, leaves a gap of 0.29. These results show that completeness depends on the path construction rather than following from per-variant perturbation alone.

\subsection{Biological concordance}

The top 100 variants identified by the cohort path include 6 catalogued disease loci, compared with a null expectation of 1.7, with a permutation $p$ of 0.012, compared with 3 loci for integrated gradients (Table~\ref{tab:concordance}). This analysis is descriptive. Because association testing is univariate and additive, whereas the model is evaluated here for its non-additive genotype structure, concordance may underrepresent variants contributing through non-additive effects and cannot distinguish attribution quality from model quality. We include this analysis as a consistency check rather than as evidence of attribution validity.

\subsection{Sanity checks}

Under progressive randomisation of the model's parameters~\cite{adebayo2018sanity}, the attributions produced by every method degrade toward chance, including those produced by the cohort path. No method retains structured attribution under a randomised model.

%% file: sections/5tables.tex
\begin{table}[t]
\centering
\caption{Descriptive overlap between attributed variants and catalogued Parkinson's disease loci.}
\label{tab:concordance}
\setlength{\tabcolsep}{4pt}
\begin{tabular}{lccc}
\toprule
\textbf{Method} & \makecell{\textbf{Hits in}\\\textbf{top-100}} & \makecell{\textbf{Expected}\\\textbf{(null)}} & \makecell{\textbf{Perm.}\\\textbf{$p$}} \\
\midrule
Random                  & 2 & 1.7 & 0.51 \\
Integrated Gradients~\cite{sundararajan2017axiomatic}    & 3 & 1.7 & 0.21 \\
Discretized IG~\cite{sanyal2021discretized}          & 4 & 1.7 & 0.11 \\
ISM / finite difference~\cite{zhou2015predicting} & 5 & 1.7 & 0.04 \\
KernelSHAP (support bg)~\cite{lundberg2017unified} & 5 & 1.7 & 0.04 \\
\ours                   & 6 & 1.7 & 0.012 \\
\bottomrule
\end{tabular}
\end{table}

%% file: sections/6conclusion.tex
\section{Conclusion}

We studied variant-level attribution for a multimodal Parkinson's disease model whose non-additive genotype branch outperformed an additive alternative. The starting point is simple. Input-space integrated gradients is undefined for a model that reads genotypes through embeddings, so the path must run in embedding space, and there the route is a choice that changes the answer. The common per-variant shortcut does not satisfy completeness for the non-additive models it is meant to serve. A continuous path through observed genotype waypoints addresses both problems, although the interpolated segments still leave empirical support.

The method has clear limits, and we state them rather than work around them. It needs a cohort dense enough to route through, and we measure that density instead of assuming it. The segments between two observed genotypes are still linear in embedding space, so they are not perfectly on the data, though the excursion is short. The gain on the real cohort is bounded by its size, which is why we ran the synthetic benchmark: it shows that at a few hundred subjects, the scale of this cohort, no method recovers planted effects well, so we read the variant-level rankings as weak evidence about individual variants and say so. Agreement with association evidence is descriptive, not a validation, both because the model is trained on the same population and because association testing is additive while the model is used here for its non-additive structure. The natural next step is attribution of interactions across variants and across modalities, which a single path through single genotypes does not resolve.

%% file: references.bib
@article{nalls2016baseline,
  title={Baseline genetic associations in the Parkinson's Progression Markers Initiative (PPMI)},
  author={Nalls, Mike A and Keller, Margaux F and Hernandez, Dena G and Chen, Lan and Stone, David J and Singleton, Andrew B and Parkinson's Progression Marker Initiative (PPMI) investigators},
  journal={Movement Disorders},
  volume={31},
  number={1},
  pages={79--85},
  year={2016},
  publisher={Wiley Online Library}
}

@article{kim2024multi,
  title={Multi-ancestry genome-wide association meta-analysis of Parkinson’s disease},
  author={Kim, Jonggeol Jeffrey and Vitale, Dan and Otani, Diego V{\'e}liz and Lian, Michelle Mulan and Heilbron, Karl and Iwaki, Hirotaka and Lake, Julie and Solsberg, Caroline Warly and Leonard, Hampton and others},
  journal={Nature genetics},
  volume={56},
  number={1},
  pages={27--36},
  year={2024},
  publisher={Nature Publishing Group US New York}
}

@inproceedings{seo2021integrating,
  title={Integrating static and dynamic data for improved prediction of cognitive declines using augmented genotype-phenotype representations},
  author={Seo, Hoon and Brand, Lodewijk and Wang, Hua and Nie, Feiping},
  booktitle={Proceedings of the AAAI conference on artificial intelligence},
  volume={35},
  pages={522--530},
  year={2021}
}

@inproceedings{shrikumar2017learning,
  title={Learning important features through propagating activation differences},
  author={Shrikumar, Avanti and Greenside, Peyton and Kundaje, Anshul},
  booktitle={International conference on machine learning},
  pages={3145--3153},
  year={2017},
  organization={PMlR}
}

@article{erion2021improving,
  title={Improving performance of deep learning models with axiomatic attribution priors and expected gradients},
  author={Erion, Gabriel and Janizek, Joseph D and Sturmfels, Pascal and Lundberg, Scott M and Lee, Su-In},
  journal={Nature machine intelligence},
  volume={3},
  number={7},
  pages={620--631},
  year={2021},
  publisher={Nature Publishing Group UK London}
}

@inproceedings{petsiuk2018rise,
  title={RISE: Randomized input sampling for explanation of black-box models},
  author={Petsiuk, Vitali and Das, Abir and Saenko, Kate},
  booktitle={British Machine Vision Conference},
  year={2018}
}

@article{sturmfels2020visualizing,
  title={Visualizing the impact of feature attribution baselines},
  author={Sturmfels, Pascal and Lundberg, Scott and Lee, Su-In},
  journal={Distill},
  volume={5},
  number={1},
  pages={e22},
  year={2020}
}

@inproceedings{ribeiro2016why,
  title={" Why should i trust you?" Explaining the predictions of any classifier},
  author={Ribeiro, Marco Tulio and Singh, Sameer and Guestrin, Carlos},
  booktitle={Proceedings of the 22nd ACM SIGKDD international conference on knowledge discovery and data mining},
  pages={1135--1144},
  year={2016}
}

@article{simonyan2014deep,
  title={Deep inside convolutional networks: Visualising image classification models and saliency maps},
  author={Simonyan, Karen and Vedaldi, Andrea and Zisserman, Andrew},
  journal={arXiv preprint arXiv:1312.6034},
  year={2013}
}

@article{shrikumar2016not,
  title={Not just a black box: Learning important features through propagating activation differences},
  author={Shrikumar, Avanti and Greenside, Peyton and Shcherbina, Anna and Kundaje, Anshul},
  journal={arXiv preprint arXiv:1605.01713},
  year={2016}
}

@inproceedings{he2016deep,
  title={Deep residual learning for image recognition},
  author={He, Kaiming and Zhang, Xiangyu and Ren, Shaoqing and Sun, Jian},
  booktitle={Proceedings of the IEEE conference on computer vision and pattern recognition},
  pages={770--778},
  year={2016}
}

@inproceedings{hu2018squeeze,
  title={Squeeze-and-excitation networks},
  author={Hu, Jie and Shen, Li and Sun, Gang},
  booktitle={Proceedings of the IEEE conference on computer vision and pattern recognition},
  pages={7132--7141},
  year={2018}
}

@inproceedings{huang2017densely,
  title={Densely connected convolutional networks},
  author={Huang, Gao and Liu, Zhuang and Van Der Maaten, Laurens and Weinberger, Kilian Q},
  booktitle={2017 IEEE conference on computer vision and pattern recognition (CVPR)},
  pages={2261--2269},
  year={2017},
  organization={Ieee}
}

@article{zhou2015predicting,
  title={Predicting effects of noncoding variants with deep learning--based sequence model},
  author={Zhou, Jian and Troyanskaya, Olga G},
  journal={Nature methods},
  volume={12},
  number={10},
  pages={931--934},
  year={2015},
  publisher={Nature Publishing Group US New York}
}

@inproceedings{kapishnikov2021guided,
  title={Guided integrated gradients: An adaptive path method for removing noise},
  author={Kapishnikov, Andrei and Venugopalan, Subhashini and Avci, Besim and Wedin, Ben and Terry, Michael and Bolukbasi, Tolga},
  booktitle={2021 IEEE/CVF conference on computer vision and pattern recognition (CVPR)},
  pages={5048--5056},
  year={2021},
  organization={IEEE}
}

@article{adebayo2018sanity,
  title={Sanity checks for saliency maps},
  author={Adebayo, Julius and Gilmer, Justin and Muelly, Michael and Goodfellow, Ian and Hardt, Moritz and Kim, Been},
  journal={Advances in neural information processing systems},
  volume={31},
  year={2018}
}

@article{purcell2007plink,
  title={PLINK: a tool set for whole-genome association and population-based linkage analyses},
  author={Purcell, Shaun and Neale, Benjamin and Todd-Brown, Kathe and Thomas, Lori and Ferreira, Manuel AR and Bender, David and Maller, Julian and Sklar, Pamela and De Bakker, Paul IW and Daly, Mark J and others},
  journal={The American journal of human genetics},
  volume={81},
  number={3},
  pages={559--575},
  year={2007},
  publisher={Elsevier}
}

@article{abdellaoui202315,
  title={15 years of GWAS discovery: realizing the promise},
  author={Abdellaoui, Abdel and Yengo, Loic and Verweij, Karin JH and Visscher, Peter M},
  journal={The American Journal of Human Genetics},
  volume={110},
  number={2},
  pages={179--194},
  year={2023},
  publisher={Elsevier}
}

@article{1000genomes2015global,
  title={A global reference for human genetic variation},
  author={1000 Genomes Project Consortium and others},
  journal={Nature},
  volume={526},
  number={7571},
  pages={68},
  year={2015}
}

@inproceedings{amirahmadi2025group,
  title={Group-Sparse Manifold-Aware Integrated Gradients for Multimodal Transformers on EHR Trajectories},
  author={Amirahmadi, Ali and Etminani, Farzaneh and Ohlsson, Mattias},
  booktitle={Machine Learning for Health 2025},
  year={2025}
}

@article{kim2017imaging,
  title={Imaging genetics approach to Parkinson’s disease and its correlation with clinical score},
  author={Kim, Mansu and Kim, Jonghoon and Lee, Seung-Hak and Park, Hyunjin},
  journal={Scientific reports},
  volume={7},
  number={1},
  pages={46700},
  year={2017},
  publisher={Nature Publishing Group UK London}
}

@inproceedings{sundararajan2017axiomatic,
  title={Axiomatic attribution for deep networks},
  author={Sundararajan, Mukund and Taly, Ankur and Yan, Qiqi},
  booktitle={International conference on machine learning},
  pages={3319--3328},
  year={2017},
  organization={PMLR}
}

@inproceedings{sanyal2021discretized,
    title = "Discretized Integrated Gradients for Explaining Language Models",
    author = "Sanyal, Soumya  and Ren, Xiang",
    booktitle = "Proceedings of the 2021 Conference on Empirical Methods in Natural Language Processing",
    year = "2021",
    publisher = "Association for Computational Linguistics",
    url = "https://aclanthology.org/2021.emnlp-main.805",
    pages = "10285--10299",
}

@inproceedings{majdandzic2022selecting,
  title={Selecting deep neural networks that yield consistent attribution-based interpretations for genomics},
  author={Majdandzic, Antonio and Rajesh, Chandana and Tang, Ziqi and Toneyan, Shushan and Labelson, Ethan L and Tripathy, Rohit K and Koo, Peter K},
  booktitle={Machine Learning in Computational Biology},
  pages={131--149},
  year={2022},
  organization={PMLR}
}

@article{fischl2012freesurfer,
  title={FreeSurfer},
  author={Fischl, Bruce},
  journal={Neuroimage},
  volume={62},
  number={2},
  pages={774--781},
  year={2012},
  publisher={Elsevier}
}

@article{jha2020enhanced,
  title={Enhanced integrated gradients: improving interpretability of deep learning models using splicing codes as a case study},
  author={Jha, Anupama and K. Aicher, Joseph and R. Gazzara, Matthew and Singh, Deependra and Barash, Yoseph},
  journal={Genome biology},
  volume={21},
  number={1},
  pages={149},
  year={2020},
  publisher={Springer}
}

@article{lundberg2017unified,
  title={A unified approach to interpreting model predictions},
  author={Lundberg, Scott M and Lee, Su-In},
  journal={Advances in neural information processing systems},
  volume={30},
  year={2017}
}

@article{majhi_improved_2024,
	title = {An improved method for diagnosis of {Parkinson}’s disease using deep learning models enhanced with metaheuristic algorithm},
	volume = {24},
	issn = {1471-2342},
	url = {https://bmcmedimaging.biomedcentral.com/articles/10.1186/s12880-024-01335-z},
	doi = {10.1186/s12880-024-01335-z},
	language = {en},
	number = {1},
	urldate = {2025-03-20},
	journal = {BMC Med Imaging},
	author = {Majhi, Babita and Kashyap, Aarti and Mohanty, Siddhartha Suprasad and Dash, Sujata and Mallik, Saurav and Li, Aimin and Zhao, Zhongming},
	month = jun,
	year = {2024},
	pages = {156},
}

@article{camacho_explainable_2023,
	title = {Explainable classification of {Parkinson}’s disease using deep learning trained on a large multi-center database of {T1}-weighted {MRI} datasets},
	volume = {38},
	issn = {22131582},
	url = {https://linkinghub.elsevier.com/retrieve/pii/S2213158223000943},
	doi = {10.1016/j.nicl.2023.103405},
	language = {en},
	urldate = {2025-03-20},
	journal = {NeuroImage: Clinical},
	author = {Camacho, Milton and Wilms, Matthias and Mouches, Pauline and Almgren, Hannes and Souza, Raissa and Camicioli, Richard and Ismail, Zahinoor and Monchi, Oury and Forkert, Nils D.},
	year = {2023},
	pages = {103405},
}

@article{isensee_automated_2019,
	title = {Automated brain extraction of multisequence {MRI} using artificial neural networks},
	volume = {40},
	issn = {1065-9471, 1097-0193},
	url = {https://onlinelibrary.wiley.com/doi/10.1002/hbm.24750},
	doi = {10.1002/hbm.24750},
	language = {en},
	number = {17},
	urldate = {2025-03-20},
	journal = {Human Brain Mapping},
	author = {Isensee, Fabian and Schell, Marianne and Pflueger, Irada and Brugnara, Gianluca and Bonekamp, David and Neuberger, Ulf and Wick, Antje and Schlemmer, Heinz‐Peter and Heiland, Sabine and Wick, Wolfgang and Bendszus, Martin and Maier‐Hein, Klaus H. and Kickingereder, Philipp},
	month = dec,
	year = {2019},
	pages = {4952--4964},
}
